\documentclass[letterpaper, 10 pt, journal, twoside]{IEEEtran}

\usepackage{amsmath,amsfonts}
\usepackage{algorithm}
\usepackage{array}
\usepackage[caption=false,font=normalsize,labelfont=sf,textfont=sf]{subfig}
\usepackage{textcomp}
\usepackage{verbatim}
\usepackage{graphicx}
\usepackage[hyphens]{url}
\usepackage{cite}

\usepackage{bm}
\usepackage[table]{xcolor}
\usepackage{hyperref}
\hypersetup{
    colorlinks=false,
    citebordercolor=white,
    linkbordercolor=white,
    urlbordercolor=cyan,
}
\usepackage{lipsum}

\usepackage{pifont}
\usepackage{amssymb}
\usepackage{nccmath}
\usepackage{comment}
\usepackage{balance}

\usepackage{multirow}
\usepackage{booktabs}
\usepackage{threeparttable}
\usepackage{siunitx}

\usepackage{tabularx}

\usepackage{dblfloatfix}
\usepackage[noend]{algpseudocode}
\usepackage[letterpaper, top=60pt, bottom=43pt, left=48pt, right=48pt]{geometry}
\usepackage{color}
\usepackage{soul}
\sethlcolor{white} 

\newcommand{\fig}[1]{Fig.~\ref{#1}}
\newcommand{\Eq}[1]{Eq.~(\ref{#1})}
\newcommand{\tab}[1]{Table~\ref{#1}}

\newcommand{\PREPRINTYEAR}{2025}
\newcommand{\PUBLISHEDIN}{IEEE Robotics and Automation Letters}
\newcommand{\DOI}{10.1109/LRA.2025.3580332} 

\usepackage[placement=top,vshift=-7]{background}
\SetBgScale{1.0}
\SetBgContents{\href{https://doi.org/\DOI}{DOI \DOI}}

\SetBgColor{black}
\SetBgAngle{0}
\SetBgOpacity{1.0}

\begin{document}

\thispagestyle{empty}
\onecolumn
{
  \topskip0pt
  \vspace*{\fill}
  \centering
  \LARGE{%
    \copyright{} \PREPRINTYEAR~\PUBLISHEDIN\\\vspace{1cm}
    Personal use of this material is permitted.
    Permission from \PUBLISHEDIN~must be obtained for all other uses, in any current or future media, including reprinting or republishing this material for advertising or promotional purposes, creating new collective works, for resale or redistribution to servers or lists, or reuse of any copyrighted component of this work in other works.}
    \vspace*{\fill}
}
\NoBgThispage
\twocolumn          	
\BgThispage

\title{\Large Tightly-Coupled LiDAR-IMU-Leg Odometry with Online \\Learned Leg Kinematics Incorporating Foot Tactile Information}

\author{Taku Okawara$^{1,2}$, Kenji Koide$^{2}$, Aoki Takanose$^{2}$, Shuji Oishi$^{2}$, \\Masashi Yokozuka$^{2}$, Kentaro Uno$^{1}$, and Kazuya Yoshida$^{1}$
\thanks{Manuscript received: February 4, 2025; Revised:
April 29, 2025; Accepted: May 29, 2025.}
\thanks{This paper was recommended for publication by Editor Olivier Stasse upon evaluation of the Associate Editor and Reviewers’ comments.}
\thanks{*This work was supported in part by and KAKENHI Grant Number 22KJ0292, a project commissioned by the New Energy and Industrial Technology Development Organization (NEDO).}
\thanks{$^{1}$Taku Okawara, Kentaro Uno, and Kazuya Yoshida are with Space Robotics Lab., Dept. Aerospace Eng., Tohoku University, Japan, {\tt\small taku.okawara@aist.go.jp}}%
\thanks{$^{2}$Taku Okawara, Kenji Koide, Aoki Takanose, Shuji Oishi, and  Masashi Yokozuka are with the Dept. of Information Technology and Human Factors, the National Institute of Advanced Industrial Science and Technology, Japan}%
}

\markboth{IEEE ROBOTICS AND AUTOMATION LETTERS. PREPRINT VERSION. ACCEPTED May, 2025}%
{Taku Okawara \MakeLowercase{\textit{et al.}}: Tightly-Coupled LiDAR-IMU-Leg Odometry with Online Learned Kinematics Incorporating Foot Tactile Info.} 

\maketitle

\begin{abstract}
  In this letter, we present tightly coupled LiDAR-IMU-leg odometry, which is robust to challenging conditions such as featureless environments and deformable terrains.
  We developed an online learning-based leg kinematics model named the \textit{neural leg kinematics model}, which incorporates tactile information (foot reaction force) to implicitly express the nonlinear dynamics between robot feet and the ground.
  Online training of this model enhances its adaptability to weight load changes of a robot (e.g., assuming delivery or transportation tasks) and terrain conditions.
  According to the \textit{neural adaptive leg odometry factor} and online uncertainty estimation of the leg kinematics model-based motion predictions, we jointly solve online training of this kinematics model and odometry estimation on a unified factor graph to retain the consistency of both.
  The proposed method was verified through real experiments using a quadruped robot in two challenging situations: 1) a sandy beach, representing an extremely featureless area with a deformable terrain, and 2) a campus, including multiple featureless areas and terrain types of asphalt, gravel (deformable terrain), and grass. 
  Experimental results showed that our odometry estimation incorporating the \textit{neural leg kinematics model} outperforms state-of-the-art works. Our project page is available for further details: \url{https://takuokawara.github.io/RAL2025_project_page/}
\end{abstract}


\begin{figure}[tb]
  \centering
  \includegraphics[width=1.0\linewidth]{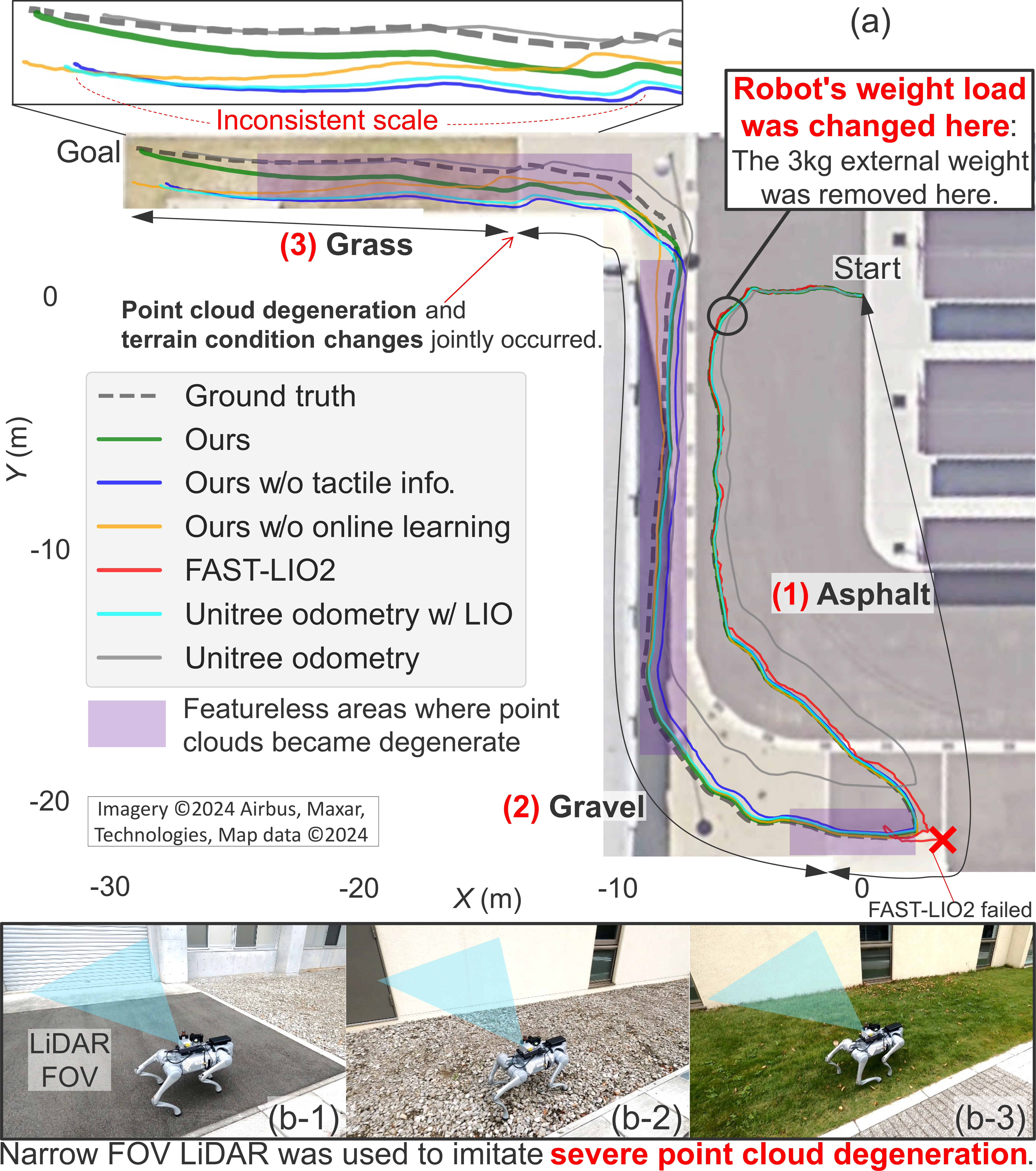}
  \caption{(a) Odometry estimation results on the campus including terrain condition changes and featureless areas (b-1, b-2, and b-3). To validate the adaptability of our online learning-based kinematics model (\textit{neural leg kinematics model}) for changes in the robot's weight load, the \SI{3}{kg} external weight was removed in the middle of this experiment.}
  \label{fig:jobutsu_traj}
\end{figure}
\section{Introduction}
\IEEEPARstart{L}{egged} robots have significant potential for transportation and inspection tasks in challenging environments (e.g., rough terrain) thanks to their superior locomotion capabilities compared to wheeled robots.
Accurate and robust odometry estimation (including SLAM) is crucial for the reliable navigation required for these tasks.
Although state-of-the-art LiDAR-IMU odometry is accurate thanks to the tight coupling of LiDAR and IMU constraints~\cite{xu2022fast}, these methods fail in featureless environments (e.g., vast flatlands such as lunar surfaces, tunnels), where LiDAR point clouds degenerate~\cite{okawara2024neuralwheel}.
IMUs can provide information on relative sensor motion by integrating their measurements independently of geometric features to mitigate estimation drift in such situations.
However, IMU-based constraints are unstable in the presence of long-term point cloud degeneration because integration errors (especially double integration of linear acceleration) accumulate rapidly due to IMU measurement noise.
For legged robots, robot motion can be estimated based on joint motions (leg forward kinematics), namely leg odometry~\cite{wisth2022vilens,yang2022online}.
For translational elements, leg forward kinematics-based constraints are more reliable than IMU-based constraints because leg kinematics-based motion prediction requires a single integration, which accumulates errors more slowly compared to double integration.
However, conventional leg odometry algorithms assume that the foot velocity with respect to the ground is zero (no foot slippage); therefore, this algorithm is also unstable under conditions such as deformable terrain and foot slippage.

To cope with challenging situations such as featureless environments and deformable terrain, we developed an odometry estimation algorithm that fuses LiDAR-IMU constraints and leg kinematics constraints in a tightly coupled way.
To address such situations where the assumptions of the conventional leg odometry algorithms become invalid, we incorporated \hl{foot tactile information (reaction force, contact state)} into the leg kinematics model to directly express the dynamic interaction between feet and the ground.
The reaction forces of feet vary depending on terrain characteristics such as soil parameters, friction coefficients, and foot sinkage~\cite{vanderkop2022novel}.
However, explicitly accurate identification of these parameters is difficult because the interaction model is complex and nonlinear.
To effectively incorporate tactile information into leg kinematics, we applied a neural network for implicitly expressing the dynamics models that are difficult to model rigorously.
We designed the network to be trained online to adapt to changes in the robot's weight load and the terrain types because the foot reaction force varies with both.
Moreover, we jointly performed odometry estimation and online training of our network on a unified factor graph to retain both consistency, based on the proposed factor named the \textit{neural adaptive leg odometry factor}.

We extended our previous work~\cite{okawara2024neuralwheel}, which developed an online learning-based wheel kinematics model without tactile information (e.g., reaction forces or torques applied to the wheels or their axes), with the following contributions:
\begin{enumerate}
  \item We developed the \textit{neural leg kinematics model}, which incorporates tactile information (foot reaction force) into the leg kinematics. This model is expressed with an online-trained network to consider its adaptability to the robot's weight load and terrain conditions.
  \item We proposed the \textit{neural adaptive leg odometry factor} to simultaneously perform odometry estimation and online training of the neural leg kinematics model on a unified factor graph to maintain their consistency. 
  \item We explicitly estimated the uncertainty (i.e., covariance matrix) of this leg kinematics-based motion constraint online to create a reasonable constraint in challenging conditions (e.g., deformable terrain).
\end{enumerate}

\section{Related works}
\subsection{Fundamental leg odometry algorithms}\label{subsec:basic_leg_odom}
Traditional leg odometry algorithms estimate robot velocity based on joint motions (leg forward kinematics) under ideal conditions where the terrain is flat and non-deformed, and none of the feet slip. 
The robot velocity $\bm{v}_b$ caused by the $j$-th leg motion is defined as follows~\cite{wisth2022vilens,yang2022online}:
\begin{align}
  \label{eq:leg_odometry}
  \bm{v}_b = - (\bm{J}_{j} ~ \dot{\bm{ \theta}}_j + \lfloor\boldsymbol{\omega}\rfloor^{\times} \text{\hl{$\rm{fk}(\bm{ \theta}_{\it{j}})$}}),
\end{align}
where 
\hl{$\rm{fk}(\bm{ \theta}_{\it{j}})$ is an operation to output the $j$-th foot position described in the robot frame through forward kinematics with its leg joint angles $\bm{ \theta}_j$}, $\dot{\bm{ \theta}}_j$ is the $j$-th leg joint angular velocities, $\bm{J}_{j}$ is the Jacobian matrix related to the $j$-th foot position and its joint angles, and $\lfloor\boldsymbol{\omega}\rfloor^{\times}$ is the skew-symmetric matrix for the angular velocity of the robot body.
To estimate $\bm{v}_b$ for robots with an arbitrary number of legs (e.g., quadrupeds or hexapods) making multiple contacts, the average linear velocity of only the feet in contact with the ground is used.
Although this leg forward kinematics-based motion prediction is reasonable under the aforementioned ideal conditions, this model is fragile under conditions with terrain deformation and foot slippage.
Michael~et~al. proposed an EKF-based approach with leg forward kinematics and IMU~\cite{bloesch2013state} to cope with such challenging situations by constraining each contact foot position with a random walk model with zero-mean Gaussian noise.

\subsection{Proprioceptive and exteroceptive sensor fusion-based odometry estimation for legged robots}\label{subsec:ep_fusion}
Fusing exteroceptive sensors (e.g., camera, LiDAR) and proprioceptive sensors is crucial for enhancing robustness of odometry estimation and SLAM algorithms to featureless environments and reducing its estimation drift.
Loosely coupled~\cite{hartley2018hybrid} and tightly coupled~\cite{wisth2019robust} visual-IMU-leg odometry based on factor graph optimization have been proposed for robustness in featureless environments with dynamic objects.
Although the above approaches successfully incorporated leg forward kinematics into exteroceptive sensor-based estimation algorithms, they did not account for errors caused by terrain-dependent phenomena (foot slippage and deformations of legs and/or the ground); thus, accurate odometry estimation is difficult in such challenging situations.
To cope with such errors, David~et~al. explicitly estimated the translational velocity bias of \Eq{eq:leg_odometry} based on a simple linear model in addition to odometry estimation~\cite{wisth2022vilens}.

Errors in leg forward kinematics significantly depend on inaccuracies in kinematic parameters (e.g., link deformation and machining errors).
In particular, a foot tip made of rubber can deform upon impact when the foot contacts hard ground; thus, the kinematic parameters of legged robots can change dynamically.
Yang~et~al. proposed performing online calibration of kinematic parameters and odometry estimation simultaneously based on a Kalman filter~\cite{yang2022online}. 

Although these approaches demonstrated accurate odometry estimation in featureless areas, their works do not assume more challenging situations, where severely featureless areas and terrain condition changes are jointly included.

\begin{figure*}[tb]
  \centering
  \includegraphics[width=1.0\linewidth]{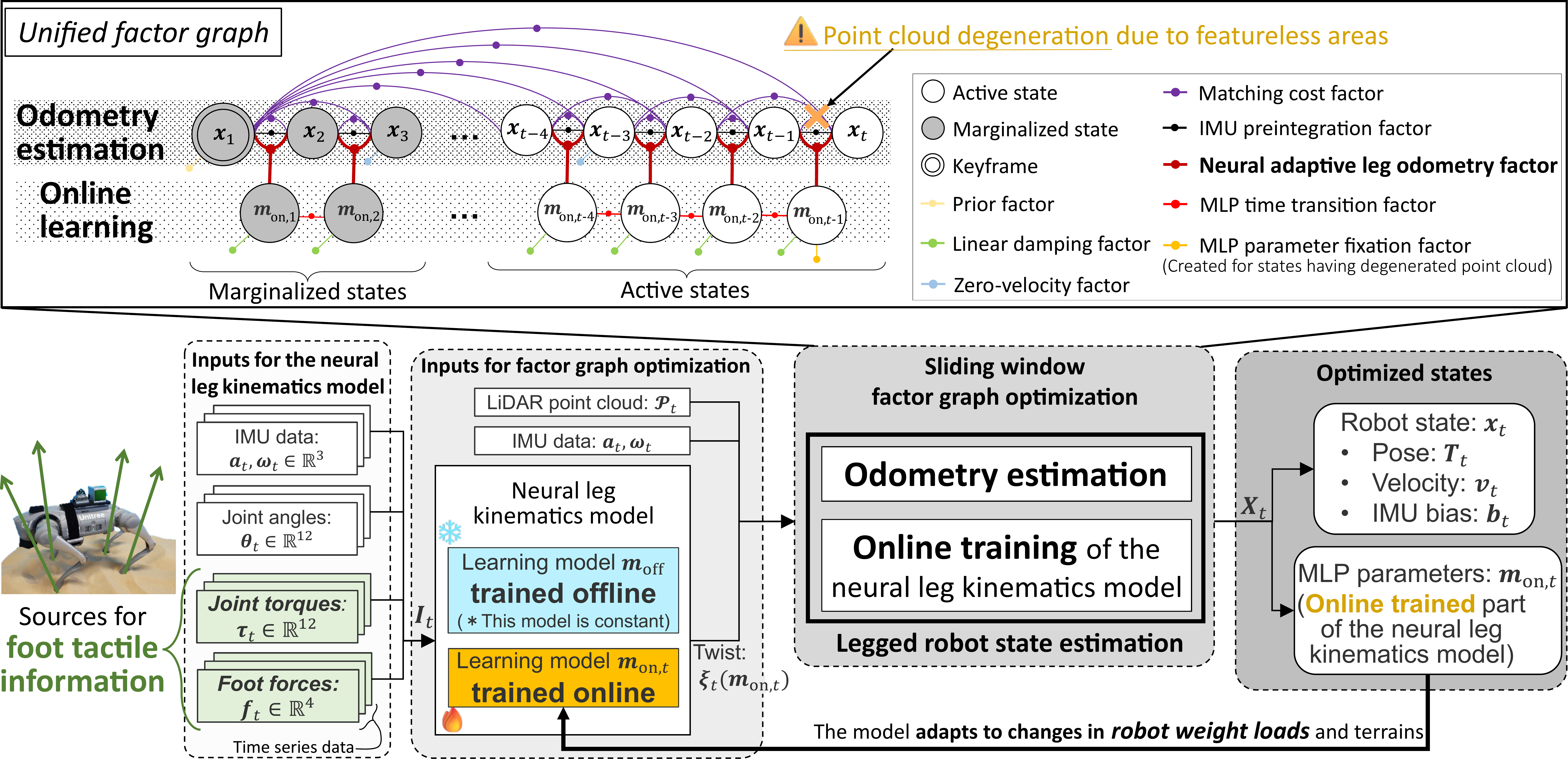}  
  \caption{Overview of the proposed framework. \hl{To accomplish reliable legged robot state estimation, we jointly perform LiDAR-IMU-leg odometry and online training of the \textit{neural leg kinematics model}, on a unified factor graph. In contrast to \mbox{\cite{okawara2024neuralwheel}}, the proposed network infers the robot twist $\bm {\xi}_t$ by incorporating foot tactile information in addition to proprioceptive sensor data.}
           }
  \label{fig:overview_chap4}
\end{figure*}
\subsection{Leg odometry model with tactile information}\label{subsec:leg_odometry_with_tactiles}
Tactile information in legged robots includes elements such as a foot reaction force and a binary contact state (whether the foot is raised or contacting the ground), which can be measured using force sensors, joint torques, or touch sensors.
Most approaches~\cite{wisth2019robust,wisth2022vilens} use these sensors to detect foot contact for leg kinematics-based motion predictions because only the feet contacting the ground influence the robot's movement through leg kinematics.
Fourmy~et~al. demonstrated that the foot contact force can be used for estimating the displacement of the body of a legged robot~\cite{fourmy2021contact}.
Specifically, the foot contact force is divided by the robot mass to estimate the acceleration caused by the force.
Finally, the velocity and displacement are estimated by integrating the acceleration; thus, these values are used as measurements for the robot motion.
Kang~et~al. extended Fourmy's work to jointly perform odometry and external force estimation on a unified factor graph~\cite{kang2023view}.

Although these studies~\cite{fourmy2021contact,kang2023view} explicitly regard the robot's mass as being constant, the mass can vary depending on robot applications (e.g., transportation tasks or adding devices for augmenting the robot's capability).

Unlike the aforementioned studies, our leg kinematics model can adapt to changes in the robot's weight-load conditions, thanks to online training of our learning model.

\subsection{Learning-based kinematic model for legged robots}
In contrast to the aforementioned approaches (Section~\ref{subsec:basic_leg_odom},\ref{subsec:ep_fusion},\ref{subsec:leg_odometry_with_tactiles}) using model-based leg forward kinematics, learning-based leg kinematics models~\cite{buchanan2022learning,wassermanlegolas} have been developed to better express terrain-dependently nonlinear dynamics such as foot slippage and deformation of legs and/or the ground.
These studies demonstrated robust and accurate odometry estimation using their learning-based kinematic model and exteroceptive sensor constraints.
A particularly relevant study is~\cite{yang2024state}, which proposes an invariant EKF-based odometry estimation based on IMU measurements and learning-based leg forward kinematics incorporating tactile information (foot reaction force).
Their learning model only outputs the weights (i.e., reliability) of each foot in multiple contacts through input data of foot force sensor values.
They only used these weights to calculate a weighted average of each leg forward kinematics-based motion for more accurate odometry estimation instead of the average described in Section~\ref{subsec:basic_leg_odom}.
They verified that the odometry estimation with learning-based leg kinematics incorporating tactile information outperformed the conventional algorithms, though their evaluation was limited to indoor flat floors where rigid foot contact is guaranteed, unlike our work.
We consider that their weights (reliability) do not work properly on deformable terrains (e.g., gravel or sandy beaches), where rigid foot contact cannot be guaranteed at any time.

These approaches~\cite{buchanan2022learning,wassermanlegolas,yang2024state} trained their learning models offline; thus, adaptability to variations in the robot’s weight load and terrain conditions was not considered.

\begin{figure*}[tb]
  \centering
  \includegraphics[width=1.0\linewidth]{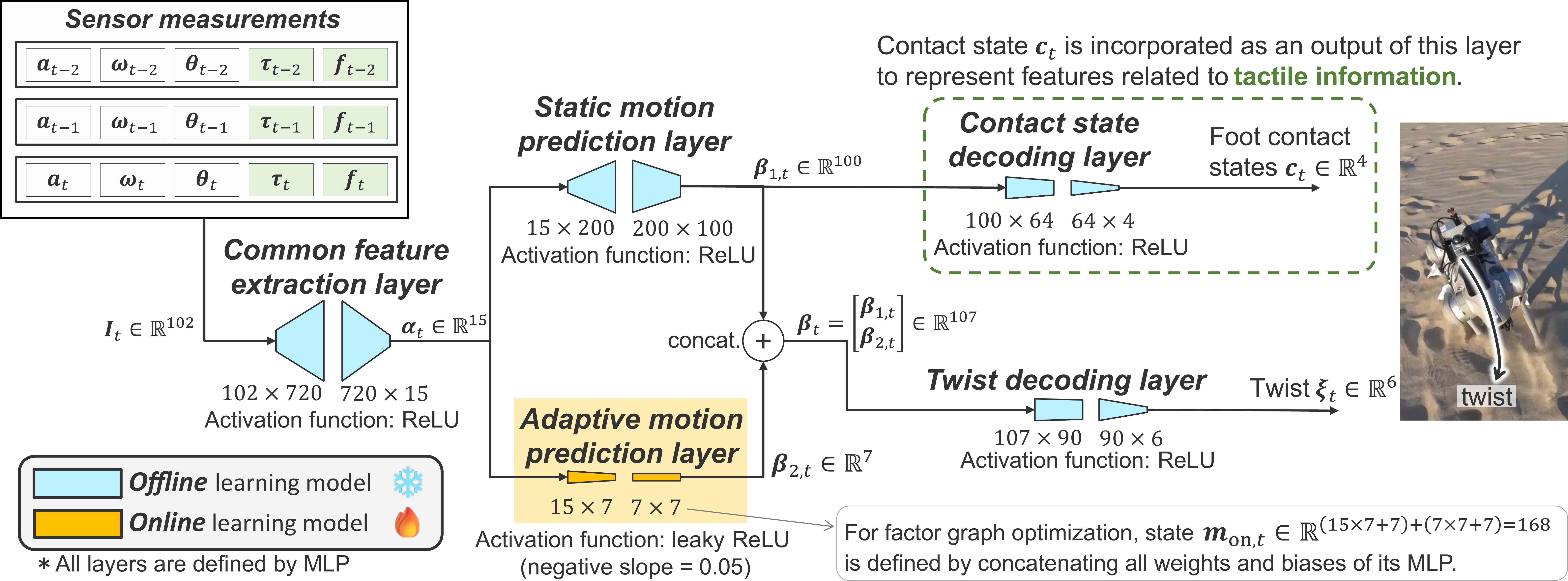}
  \caption{Neural leg kinematics model structure. This network outputs a legged robot's twist $\bm {\xi}_t \in ~\mathbb{R}^6$ and the Boolean contact states for each foot $\bm {c}_t \in ~\mathbb{R}^4$ by inputting $\bm {I}_t$.
  The dimension of the online learning model's MLP parameter $\bm{m}_{\text{on},t}$ is 168.
           }
  \label{fig:prop_odom_net}
\end{figure*}
\section{Methodology overview}
\subsection{System overview}
As shown in \fig{fig:overview_chap4}, we jointly perform odometry estimation and online training of the \textit{neural leg kinematics model} on a unified factor graph to maintain the consistency of all constraints.
Online training of this model enhances adaptability to the weight loads of a legged robot and terrain conditions, enabling effective use of foot tactile information (reaction forces) for motion prediction because foot reaction forces vary with both robot weight loads and terrain types.
To balance accuracy and computational costs for training the network, we divided the model into two models trained \textit{online} and \textit{offline}~\cite{okawara2024neuralwheel}.
The model trained offline is constant during odometry estimation.
The robot kinematic state ${\bm X}_t$ to be estimated at time $t$ is defined as follows:
\begin{align}
  \label{eq:state_declation_chap4}
  {\bm X}_t = [{\bm x}_t, {\bm{m}}_{\text{on},t}], \\
  {\bm x}_t = [{\bm T}_t, {\bm v}_t, {\bm b}_t], 
\end{align}
where ${\bm x}_t$ is the robot state at time $t$; ${\bm T}_t~=~[{\bm R}_t | {\bm t}_t]~\in~SE(3)$ and ${\bm v}_t~\in~\mathbb{R}^3$ are the robot pose and linear velocity in the world frame, respectively;  ${\bm b}_t~=~[{\bm b}_t^a, {\bm b}_t^{\omega}]~\in~\mathbb{R}^6$ is the bias of the IMU linear acceleration ${\bm a}_t$ and the angular velocity ${\bm \omega}_t$; and ${\bm{m}}_{\text{on},t}$ is the weight and bias of the MLP of the neural leg kinematics model trained online. 
${\bm{m}}_{\text{on},t}$ is dynamically updated along with ${\bm x}_t$ with the assumption that the robot walks in a feature-rich environment at the beginning of the estimation. 
Once the online-trained network converges to a state adapted to the current terrain and robot weight load, the neural network-based motion constraint becomes reliable for optimization and enables accurate odometry estimation even in such severe conditions.
Note that \hl{the proposed method constantly trains our network online along with odometry estimation, based on factor graph optimization} to adapt to dynamic changes in weight load and terrain conditions.

\subsection{Overview of neural leg kinematics model}
\textbf{Network input and output: }
We developed a neural network (\textit{neural leg kinematics model}) that outputs the 6DOF twist (translational and rotational velocities in the tangent space of the robot body) with proprioceptive sensors (joint encoder and IMU) and tactile information (foot reaction force).
The input data for the network are time series data consisting of IMU data (linear acceleration~${\bm a}_t~\in~\mathbb{R}^3$, angular velocity~${\bm \omega}_t~\in~\mathbb{R}^3$), joint angles~${\bm \theta}_t~\in~\mathbb{R}^{12}$, joint torques~${\bm \tau }_t~\in~\mathbb{R}^{12}$, and foot force sensor values~${\bm f}_t = [{}^{\text{LF}}f_t~~{}^{\text{LH}}f_t~~{}^{\text{RH}}f_t~~{}^{\text{RF}}f_t] \in \mathbb{R}^4$.
We define $\bm {i}_t~\in~\mathbb{R}^{34}$ by concatenating the above sensor data as follows:
\begin{align}
  \label{eq:input_new_Leg_i}
  \bm {i}_{t} = [{\bm a}_t^\top~~{\bm \omega}_t^\top~~{\bm \theta}_t^\top~~{\bm \tau}_t^\top~~{\bm f}_t^\top]^\top.
\end{align}
We concatenate $\bm {i}_t$ to create the time series data of sensor values with a window size $N_{\rm w}$.
\hl{The model can implicitly represent various physical quantities (e.g., joint angular velocity based on the difference of consecutive time of joint angles) by using the time series data as inputs.}
Finally, the input for the network is defined as $\bm {I}_t = \rho([ \bm {i}_t, \bm {i}_{t-1}, ... \bm {i}_{t-{N_{\rm {w}}}-1}])$, where $\rho$ is a standardization operation.
In our implementation, we set $N_{\rm w}$ to 3; thus, the dimension of $\bm {I}_t$ is 102 ($34\times 3$).

We propose incorporating tactile information into leg kinematics models to enable robust motion prediction for terrain-dependent phenomena (e.g., terrain deformation, foot slippage).
Foot tactile information (reaction forces) can address these severe conditions because the reaction forces are influenced by foot slippage (friction forces) and terrain deformation~\cite{vanderkop2022novel}. 
In other words, foot reaction forces contain information on terrain-dependent features. 
As stated in Section~\ref{subsec:leg_odometry_with_tactiles}, 3D foot reaction forces can also be used to express a robot's velocity.
Because 3D foot reaction force measurements are not directly available on our robot platform (Unitree Go2), joint torques are used as the input data instead.
The reason is that foot reaction forces can be expressed based on the motion equation for legged robots using joint torques~\cite{kang2023view}.
Because the accuracy of motion-equation-based reaction force expression depends on a precise model identification for such as unknown disturbances and robot's mass, inertia moment~\cite{kang2023view}, learning-based approaches are preferred to accurately express such complex terms by implicit representation~\cite{hu2017contact}.
In addition, a 1D foot force sensor (Unitree Foot Pad) is used to complement the information on the 3D reaction force represented by the joint torque.

\begin{figure}[tb]
  \centering
  \includegraphics[width=1\linewidth]{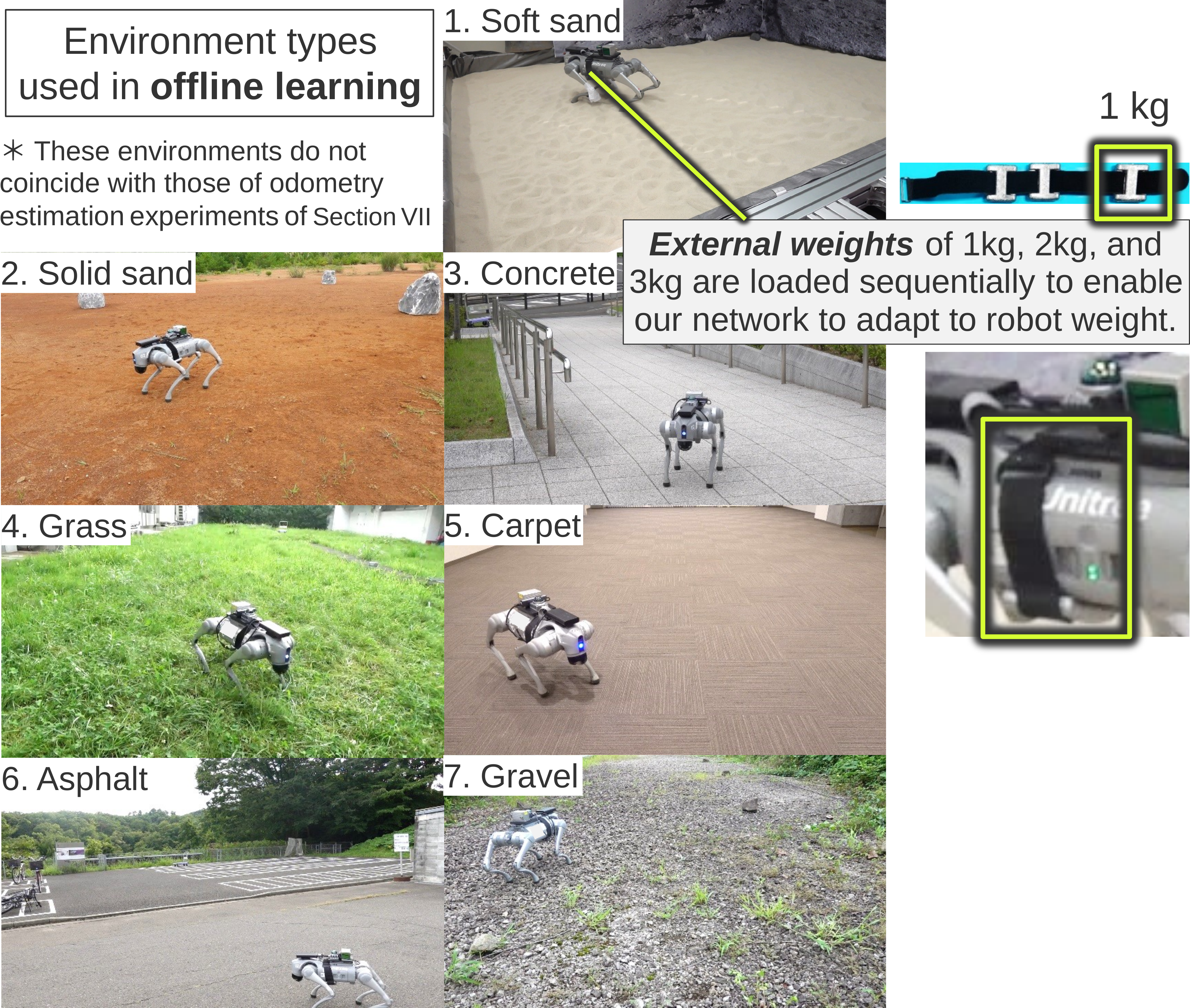}
  \caption{Datasets of each terrain for offline batch learning.}
  \label{fig:dataset_go2}
\end{figure}
\textbf{Network structure: }
As shown in Fig.~\ref{fig:prop_odom_net}, the network structure is divided into an \textit{offline learning model} and \textit{online learning model} based on our previous work work~\cite{okawara2024neuralwheel}, which developed the online trained wheel odometry model without force or torque of wheels.
Our online learning model expresses dynamic features regarding terrains and legged robots' weight loads.
Whereas, our offline learning model captures features invariant to weight and terrain, such as basic leg forward kinematics and inertial propagation.
We designed the above network structure to balance the computational cost and accuracy for realizing reasonable online learning.
The size of the online learning model is smaller than that of the offline learning model because features invariant to weight loads and terrains have a greater influence than dynamic features varying with these conditions in leg kinematics models.

First, the \textit{common feature extraction layer} takes $\bm{I}_t$ as input and extracts common features $\bm{\alpha}_t$ related to motion prediction.
The subsequent \textit{static} and \textit{adaptive motion prediction layers} take these common features $\bm{\alpha}_t$ as input and output feature vectors $\bm{\beta}_{\text{1},t}$ and $\bm{\beta}_{\text{2},t}$ to predict the robot twist in a latent feature space, respectively.
$\bm{\beta}_{\text{1},t}$ also includes features related to foot contact states to account for foot contact effects in our learning model. 
The \textit{contact state decoding layer} interprets $\bm{\beta}_{\text{1},t}$ to determine whether each foot is in contact with the ground. 
The contact state $\bm{c}_{t}$ is defined on a scale from 0 to 1, where 0 and 1 indicate non-contact and contact, respectively.
Note that $\bm{c}_{t}$ is incorporated as an output of this model to represent features related to tactile information; however, it is not directly used for online learning (factor graph optimization).
Finally, the \textit{twist decoding layer} decodes the latent features $\bm{\beta}_{t}$ into the twist in the robot body $\bm {\xi}_t$.

\section{Training the offline learning model}\label{chap4_offline}
\textbf{Training procedure: }
The offline learning model is trained to describe features invariant to the robot's weight loads and terrains for accurate motion prediction.
The offline batch learning is conducted under the condition that, while the online learning model (adaptive motion prediction layer) varies depending on the robot's weight loads and terrain, the offline learning model is common for these conditions.
By including diverse environments and the robot's weight load configurations in datasets for the offline learning process (\fig{fig:dataset_go2}), we ensure that the offline learning model provides the invariant features that can complement the adaptive responses of the online learning model under varying conditions. 
This structure aims to improve the performance and adaptability of our model across different terrains and payload scenarios.
To meet the aforementioned training conditions, we minimize the overall loss function~$\mathcal{L}$ (\Eq{eq:all_loss2}), which is defined by the sum of the loss function $\mathcal{L}_{ n}(\bm{m}_\text{off},\bm{m}_{\text{on},n})$ for each dataset.
Offline batch learning is conducted by minimizing $\mathcal{L}$ such that the MLP parameters for the offline learning model $\bm{m}_\text{off}$ are common for all datasets.
In contrast, the MLP parameters for the online learning model $\bm{m}_{\text{on},n}$ vary depending on the type of dataset.
\begin{align}
  \label{eq:all_loss2}
  \mathcal{L} &= \sum_{n=1}^{N_{\rm s}} \mathcal{L}_{ n}(\bm{m}_\text{off},\bm{m}_{\text{on},n}),\\
  \label{eq:each_loss2}
  \mathcal{L}_{ n}(\bm{m}_\text{off},\bm{m}_{\text{on},n}) &=  e_{t,\bm {\xi}} + w_1 e_{t,\bm {c}} + w_2 e_{t,\rm {r}},\\
  \label{eq:twist_loss}
  e_{t,\bm {\xi}} &= \frac{1}{n_{\rm T}} \sum_{t=1}^{n_{\rm T}} (e_{t, \rm {trans}} + w_3 e_{t, \rm {rot}}).
\end{align}
$e_{t,\bm {\xi}}$, $e_{t,\bm {c}}$, and $e_{t,\rm {r}}$ are the loss for the twist (mean square error), contact states of each foot (binary cross entropy), and $\rm{L}_2$ regularization terms for features $\bm{\beta}_{\text{2},t}$ of the online learning model, respectively. $n_{\rm T}$ is the number of training data in each dataset; $e_{t, \rm {trans}}$ and $e_{t, \rm {rot}}$ are the square error in the translation and rotation components of the twist $\bm {\xi}_t$, respectively. $w_1$~(e.g., 5), $w_2$~(e.g., $10^{-3}$), and $w_3$~(e.g., 200) are weights adjusting the scales of the loss values.

\textbf{Offline training details: }
We used $N_{\rm{s}}$ sequences including $N_{\rm{e}}$ environment types and $N_{\rm{m}}$ robot weight loads for the offline training.
$N_{\rm{e}}$ and $N_{\rm{m}}$ were set to 7 and 4, respectively, resulting in $N_{\rm{s}} = 28$. 
We changed the weight load by equipping the robot with external weights of \SI{1}{kg}, \SI{2}{kg}, and \SI{3}{kg}.
The duration of each sequence was about 10 minutes.
As shown in \fig{fig:dataset_go2}, we included a variety of environments, such as deformable and solid terrains as well as slopes. 

We used the Adam optimizer~\cite{KingBa15} with a $5\times 10^{-4}$ learning rate, 2500 epochs, and a batch size of 50 in our implementation.
The number of layers and activation functions for each MLP are shown in \fig{fig:prop_odom_net}.
\hl{We determined the reference of contact states $\bm{c}_{t}$ by offline analysis based on thresholding of \mbox{each foot’s force sensor values}, similar to related work~\mbox{\cite{wisth2022vilens}}.}
We used LiDAR-IMU odometry with an omnidirectional FOV LiDAR (Livox MID-360) to obtain reference data for the twist $\bm {\xi}_{t, \rm{L}}$.
We calculate $e_{t, \rm {trans}}$ and $e_{t, \rm {rot}}$ in \Eq{eq:twist_loss} based on the error between $\bm {\xi}_{t, \rm{L}}$ and $\bm {\xi}_{t}$ inferred by the network.
LiDAR-IMU odometry is highly accurate in feature-rich environments (e.g., our datasets in \fig{fig:dataset_go2}).
The robot also moved locally in each environment to always remain on the local map; thus, accumulation errors did not occur in LiDAR-IMU odometry for the offline training sequences.
We manually operated the legged robot (Unitree Go2) through various motions (e.g., straight and curved trajectories on slopes and steps).

\hl{\textbf{Network evaluations: }  
To demonstrate the importance of incorporating tactile information and assigning a large rotational weight (e.g., $w_3 = 200$) in the loss function~\mbox{\Eq{eq:twist_loss}}, we compared the proposed network with 1) \textit{network w/o tactile information}, and 2) \textit{network w/ small rotational weight} ($w_3=1$).}
\hl{\mbox{\tab{tab:RTE_of_network}} shows the average of relative trajectory errors (RTEs) per \mbox{\SI{1}{m}} of the proposed network and the other networks using $10\%$ of the offline training datasets (\mbox{\fig{fig:dataset_go2}}) as validation data.}
\begin{table}[t]
  \centering
  \caption{\hl{Translational and rotational RTEs per \mbox{\SI{1}{m}} of the proposed network and ablation cases.}}
  \label{tab:RTE_of_network}
  \begin{threeparttable}
    \setlength{\tabcolsep}{1.0mm}
    \begin{tabular}{cc|cc}
      \hline
      \multicolumn{2}{c|}{Method} & $t_{\text{RTE}}$~[\SI{}{m}] & \cellcolor[HTML]{EFEFEF}$r_{\text{RTE}}$~[\SI{}{deg}] \\ \hline \hline
      
      \multicolumn{2}{c|}{Proposed network (w/ tactile info., $w_3=200$)} & \textbf{0.06}  $\pm$ 0.04 & \cellcolor[HTML]{EFEFEF}\textbf{1.00}  $\pm$ 0.69 \\
      
      \multicolumn{2}{c|}{Network w/o tactile information} & 0.10  $\pm$ 0.07 & \cellcolor[HTML]{EFEFEF}1.06  $\pm$ 0.69 \\

      \multicolumn{2}{c|}{Network w/ small rotational weight ($w_3=1$)} & 0.07  $\pm$ 0.04 & \cellcolor[HTML]{EFEFEF}3.36  $\pm$ 2.87 \\

      \hline
    \end{tabular}
  \end{threeparttable}
\end{table}
\hl{For translational RTEs $t_{\text{RTE}}$, the proposed network (\mbox{\SI{0.06}{m}}) is approximately twice as accurate as the network without tactile information (\mbox{\SI{0.10}{m}}).
Regarding rotational RTEs $r_{\text{RTE}}$, the proposed network (\mbox{\SI{1.00}{deg}}) is three times more accurate than the network with a small rotational weight (\mbox{\SI{3.36}{deg}}).
The result demonstrated 1) the importance of tactile information for accurate motion prediction in challenging terrains and 2) the effectiveness of the large rotational weight $w_3$ for loss~\mbox{\Eq{eq:twist_loss}} in the proposed network.}

\section{Training online learning model}
\subsection{Neural adaptive leg odometry factor}
We train the online learning model (adaptive motion prediction layer) on-the-fly with the fixed offline learning model.
To perform online training of the network and odometry estimation on a unified factor graph, we propose the \textit{neural adaptive leg odometry factor}, which accounts for both constraints regarding robot motion and online learning.
In the online learning phase, the parameter vector $\bm{m}_{\text{on}}$ concatenating the weights and biases of MLPs for the online learning model is optimized such that online trained leg kinematics-based, LiDAR-based, and IMU-based constraints become consistent.

To derive the error in the neural adaptive leg odometry factor, we first calculate the relative pose by integrating the corresponding twist $\bm {\xi}_{i}$ of the robot with a time step $\Delta t_{i}$.
Then, we map the relative pose from $se(3)$ to $SE(3)$ space based on exponential map operation (i.e., $\rm{exp}$$(\bm {\xi}_{i} \Delta \it{t_{i}})$).
Finally, the error in this factor is defined as follows:
\begin{align}
  \label{eq:legOdomError}
  e^{\rm {Leg}}({{\bm T}_{i-1}, {\bm T}_i, {\bm{m}}_{\text{on},i}}) &= ({{\bm r}_{i}^{\rm {Leg}}})^\top ({{\bm C}_{i}^{\rm {Leg}}})^{-1} {\bm r}_{i}^{\rm {Leg}}, \\
  \label{eq:legOdomResidual}
  \bm r_{i}^{\rm {Leg}} &= \log ( {{\bm T}_{i-1}^{-1} {\bm T}_i \exp (\bm {\xi}_{i} \Delta t_{i}) ^{-1}} ),\\
  \bm {\xi}_i &= \text{NN} (\bm{m}_{\text{on},i}),
\end{align}
where $({{\bm C}_{i}^{\rm {Leg}}})^{-1}$ is the covariance matrix representing the uncertainty in the network-based motion prediction.
\hl{$\text{NN}$ is an operation to output $\bm {\xi}_i$ through the neural leg kinematics model, which incorporates the online optimized MLP parameters $\bm{m}_{\text{on},i}$, fixed offline trained model~$\bm{m}_{\text{off}}$, and the model input (i.e., time series sensor data $\bm{I}_i$), as seen in \mbox{\fig{fig:overview_chap4}}.}
\hl{We compute the Jacobian matrix related to $\bm r_{i}^{\rm {Leg}}$ and $\bm{m}_{\text{on},i}$ by using the LibTorch library to take advantage of automatic differentiation wrapped the results with a GTSAM nonlinear factor class.}

\subsection{Online uncertainty estimation for neural adaptive leg odometry factor}
We explicitly estimate the uncertainty (i.e., ${{\bm C}_{i}^{\rm {Leg}}}$) associated with the \textit{neural adaptive leg odometry factor} online, ensuring that this constraint remains reasonable for challenging terrains (e.g., rough and deformable surfaces).
We estimate ${{\bm C}_{i}^{\rm {Leg}}}$ by analyzing $\bm r_{i}^{\rm {Leg}}$ (\Eq{eq:legOdomResidual}), which is the residual of the network-based motion prediction.
$N$ (e.g., 15) samples of $\bm r_{i}^{\rm {Leg}}$ (i.e., $\bm r_{i}^{\rm {Leg}}$,...,$\bm r_{i-N}^{\rm {Leg}}$) are used to properly capture the current terrain condition with the assumption that $\bm{m}_{\text{on},t}$ does not change drastically across the $N$ samples.
\hl{The sliding window adjusts a balance between stability and responsiveness for estimating the reasonable uncertainty.}
According to the definition of variance, for example, the variance of translational X element ${\sigma_{i,\rm{Tx}}}^2$ is estimated as follows:
\begin{align}
    \label{eq:var}
    {\sigma_{i,\rm{Tx}}}^2 &= \frac{1}{N - 1} \sum_{j=1}^N r_{i,\rm{Tx}, \it{j}}^2.
\end{align}
The other five variances can be estimated similarly to that for the translational X element.
When we assume that each element of $\bm r_{i}^{\rm {Leg}}$ is independent, the covariance matrix ${{\bm C}_{i}^{\rm {Leg}}}$ can be obtained by setting these variances to the diagonal elements of ${{\bm C}_{i}^{\rm {Leg}}}$.
In addition, we assume that the estimation of poses is accurate in elements where point clouds do not degenerate; thus, we consider $\bm r_{i}^{\rm {Leg}}$ for the corresponding elements to be reliable for estimating variances.

\section{Implementation details}\label{Implementation2}

\hl{The objective function (i.e., factor graph of \mbox{\fig{fig:overview_chap4}}) for the optimization is the sum of the LiDAR-based (matching cost factor~\mbox{\cite{koide2021globally}}), IMU-based (IMU pre-integration factor~\mbox{\cite{forster2016manifold}}), leg kinematics model-based constraints, and others (MLP time transition factor, MLP parameter fixation factor~\mbox{\cite{okawara2024neuralwheel}}, and prior factor).}
\hl{Regarding factors other than the neural adaptive leg odometry factor, overviews and error functions are described in~\mbox{\cite{okawara2024neuralwheel}}}.
We implemented the factor graph using \href{https://github.com/borglab/gtsam}{GTSAM} and incrementally optimized this graph with iSAM2~\cite{kaess2012isam2}.
We used a fixed lag smoother to ensure real-time optimization within a time window (e.g.,~\SI{6}{s}).
Only active states in the time window are optimized, while marginalized states are fixed.

\begin{figure}[tb]
  \centering
  \includegraphics[width=1.0\linewidth]{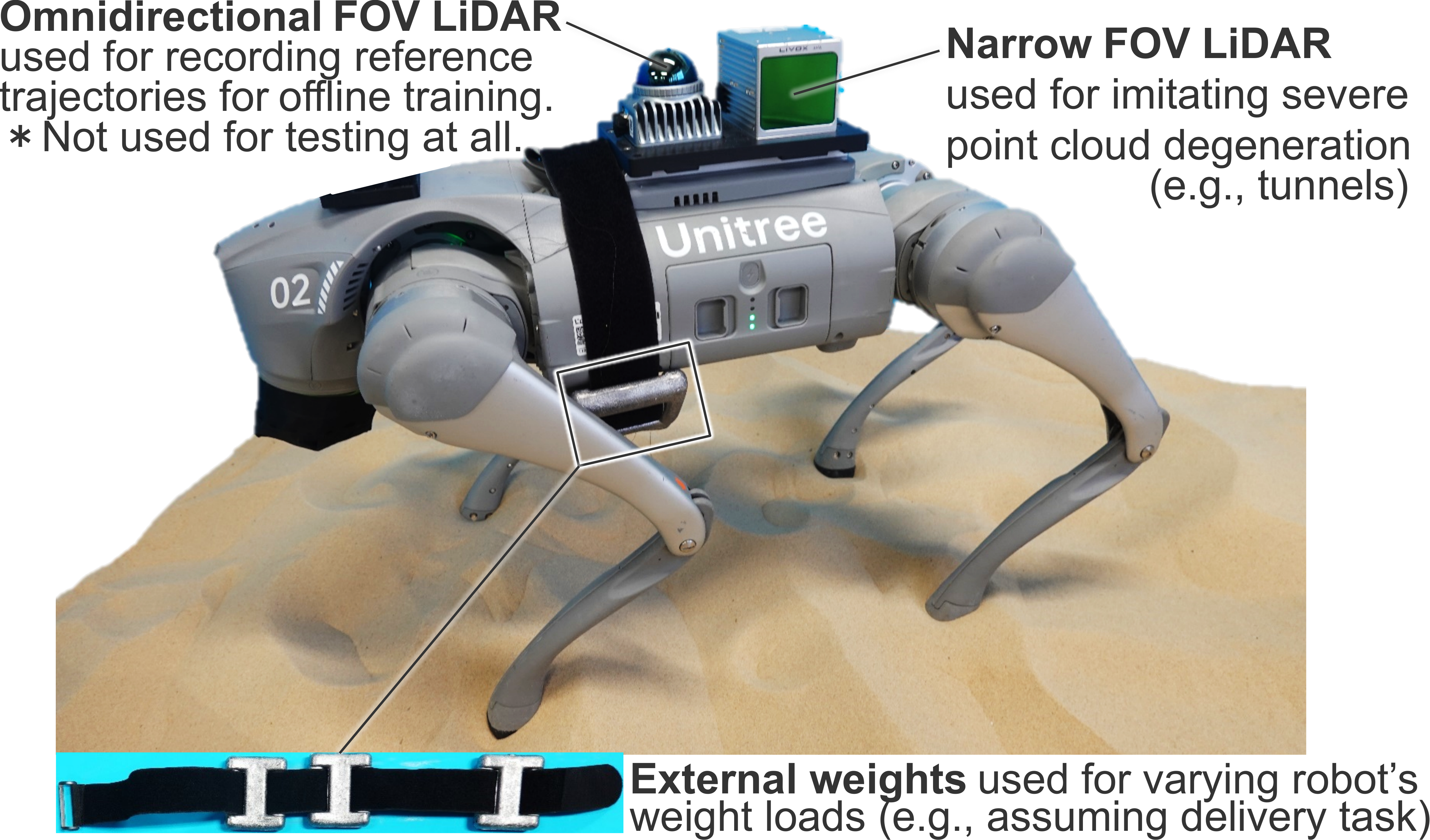}
  \caption{Unitree Go2 used as a testbed. Note that the omnidirectional LiDAR was used for recording reference trajectories for offline training, and experiments (\fig{fig:jobutsu_traj}, \fig{fig:sunahama_traj}) were conducted using only the narrow FOV LiDAR to imitate severe point cloud degeneration (e.g., tunnels and lunar surfaces).}
  \label{fig:testbed_go2}
\end{figure}

\section{Experimental results}\label{chap4}

\textbf{Experimental conditions: }
We conducted experiments with a Unitree Go2 (\fig{fig:testbed_go2}) equipped with a narrow FOV LiDAR (Livox AVIA) and external weights as a testbed to verify the proposed odometry algorithm.
We recorded point clouds, IMU data, joint angles, joint torques, and foot forces at frequencies of \SI{10}{Hz}, \SI{60}{Hz}, \SI{500}{Hz}, \SI{500}{Hz}, and \SI{500}{Hz}, respectively, to estimate robot states $\bm{x}_t$ and MLP parameters $\bm{m}_{\text{on},t}$.
We obtained the ground truth trajectories using a total station (Leica TS16).
We compared the proposed method (ours) with the following baseline methods:
\begin{itemize}
  \item \textit{Ours w/o online learning}: We created an ablative method by incorporating the fixed neural network-based motion constraints into the factor graph instead of the \textit{neural adaptive leg odometry factor}, like related works~\cite{buchanan2022learning,wassermanlegolas}. The network was trained by simple offline batch learning with all datasets (\fig{fig:dataset_go2}); thus, the \textit{adaptive motion prediction layer} (\fig{fig:prop_odom_net}) was not trained online.
  \item \textit{Ours w/o tactile information}: A network without tactile information (joint torques~${\bm \tau }_t~\in~\mathbb{R}^{12}$ and foot force sensor values~${\bm f}_t~\in~\mathbb{R}^4$) was used as a second ablative method to create motion constraints instead of the neural adaptive leg odometry factor, similar to~\cite{okawara2024neuralwheel,wisth2022vilens}.
  \item \textit{FAST-LIO2}: FAST-LIO2 is state-of-the-art odometry based on tightly-coupled LiDAR-IMU constraints~\cite{xu2022fast}.
  \item \textit{Unitree odometry}: Go2 outputs proprioceptive sensor-based odometry by a Unitree proprietary algorithm.
  \item \textit{Unitree odometry w/ LIO}: Matching cost factor~\cite{koide2021globally}, IMU pre-integration factor~\cite{forster2016manifold}, and the above Unitree odometry algorithm-based motion constraint are incorporated into a factor graph for reasonable comparison.
\end{itemize}

\begin{table}[t]
  \scriptsize
  \centering
  \caption{ATEs and RTEs of the odometry algorithms.}

  \label{tab:ATE_and_RTE}
  \begin{threeparttable}
    \setlength{\tabcolsep}{1.05mm}
    \begin{tabular}{cc|cc|cc}
      \hline
      \multicolumn{2}{c|}{\multirow{-0.5}{*}{Method/Sequence}} & \multicolumn{2}{c|}{Campus} & \multicolumn{2}{c}{Sandy beach} \\ \cline{3-6}
      
      \multicolumn{2}{c|}{\multirow{-2}{*}{}} & ATE~[\SI{}{m}] & \cellcolor[HTML]{EFEFEF}RTE~[\SI{}{m}] & ATE~[\SI{}{m}] & \cellcolor[HTML]{EFEFEF}RTE~[\SI{}{m}] \\ \hline \hline
      
      \multicolumn{2}{c|}{Ours} & 
      \textbf{0.29} $\pm$ 0.12 &     \cellcolor[HTML]{EFEFEF}{\textbf{0.13}}  $\pm$ 0.04 & 
      \textbf{0.08} $\pm$ 0.05 &     \cellcolor[HTML]{EFEFEF}{\textbf{0.12}}  $\pm$ 0.04 \\ 
      
      \multicolumn{2}{c|}{Ours w/o online learning} & 
      0.36 $\pm$ 0.17 & \cellcolor[HTML]{EFEFEF}{0.17}  $\pm$ 0.07 &   
      0.90 $\pm$ 0.70 & \cellcolor[HTML]{EFEFEF}{0.20}  $\pm$ 0.10 \\  

      \multicolumn{2}{c|}{Ours w/o tactile info.} &       
      0.63 $\pm$ 0.30 & \cellcolor[HTML]{EFEFEF}{0.15}  $\pm$ 0.06 &   
      0.12 $\pm$ 0.06 & \cellcolor[HTML]{EFEFEF}{0.12}  $\pm$ 0.04 \\  

      \multicolumn{2}{c|}{FAST-LIO2} &       
      Corrupted & \cellcolor[HTML]{EFEFEF}{Corrupted} &    
      Corrupted & \cellcolor[HTML]{EFEFEF}{Corrupted} \\  

      \multicolumn{2}{c|}{Unitree odometry w/ LIO} & 
      0.57 $\pm$ 0.32 & \cellcolor[HTML]{EFEFEF}{0.15}  $\pm$ 0.06 &   
      No record & \cellcolor[HTML]{EFEFEF}{No record} \\  

      \multicolumn{2}{c|}{Unitree odometry} & 
      0.80 $\pm$ 0.46 & \cellcolor[HTML]{EFEFEF}{0.17}  $\pm$ 0.06 &   
      No record & \cellcolor[HTML]{EFEFEF}{No record} \\  

      \hline

    \end{tabular}

  \end{threeparttable}

\end{table}
\begin{figure}[tb]
  \centering
  \includegraphics[width=1\linewidth]{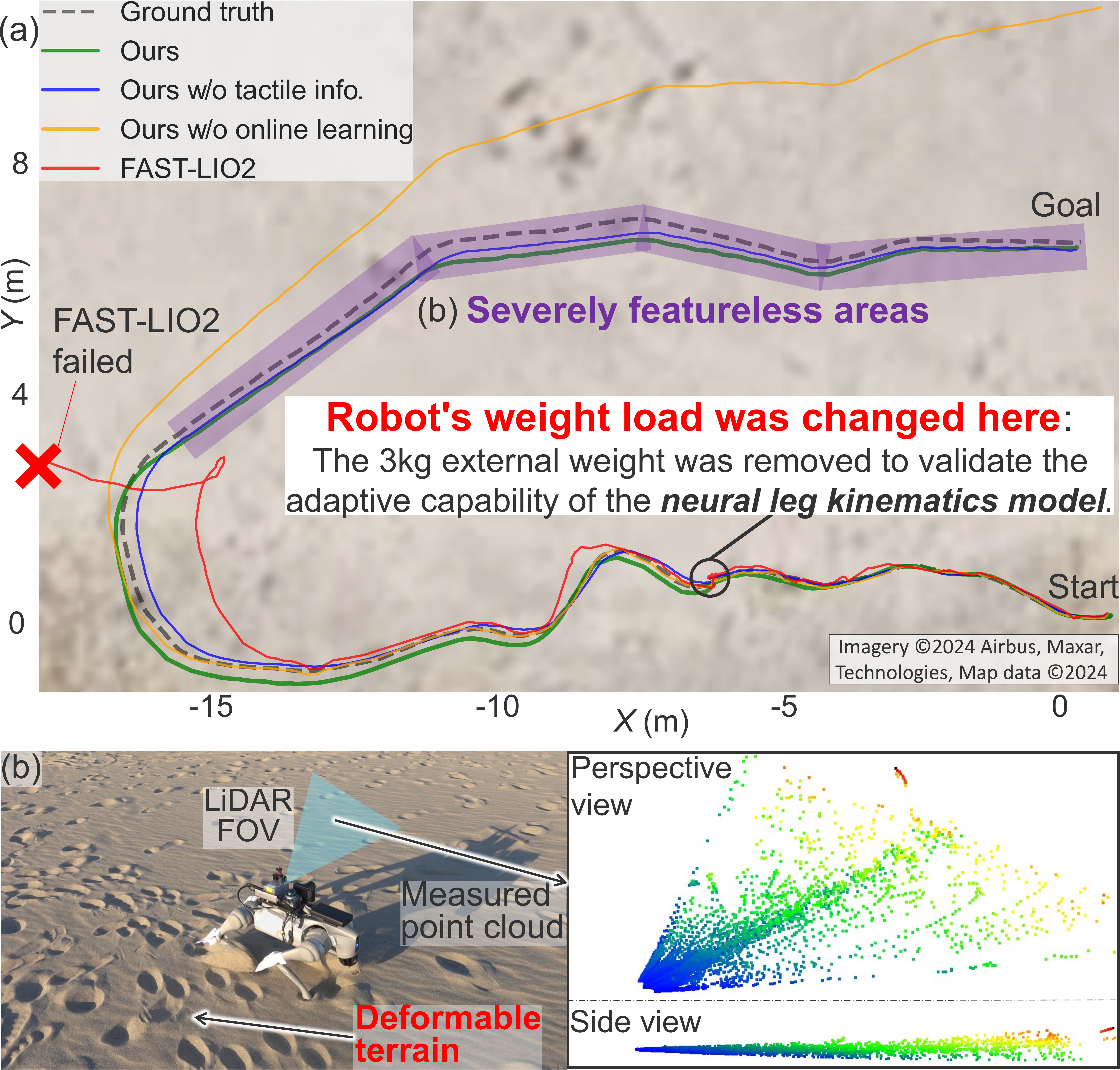}
  \caption{(a) Odometry estimation results in the sandy beach, including (b) severely featureless areas.}
  \label{fig:sunahama_traj}
\end{figure}

\textbf{Accuracy evaluation: }
We conducted experiments in two situations: 1) a campus having featureless areas (\fig{fig:jobutsu_traj}(b-1,2,3)) and terrain condition changes, and 2) a sandy beach that included severely featureless areas (\fig{fig:sunahama_traj}(b)).
\hl{Note that these environments were not included in the offline training datasets (\mbox{\fig{fig:dataset_go2}}).}
\hl{To demonstrate robustness against rapid movements, we manually operated the testbed at its maximum walking speed (approximately \mbox{\SI{2}{m/s}}), and the speed of the robot naturally varied due to acceleration, deceleration, and ground surface changes.}
\hl{To initialize the online learning model, we used \mbox{$\bm{m}_{\text{on}}$} obtained from the offline learning phase under the condition where the terrain type was grass and no additional weight was applied, to show the adaptability of online learning.
We consider that the best initial values of this parameter can be set by combining terrain classification algorithms if the proposed method is deployed in a real operation.}

\fig{fig:jobutsu_traj}(a) and \fig{fig:sunahama_traj}(a) show the trajectory comparison results.
\tab{tab:ATE_and_RTE} also shows the absolute trajectory errors (ATEs) and relative trajectory errors (RTEs)~\cite{zhang2018tutorial} for all methods.
According to these trajectory estimation results and our method's ATEs (campus: \SI{0.29}{m}; sandy beach: \SI{0.08}{m}) and RTEs (campus: \SI{0.13}{m}; sandy beach: \SI{0.12}{m}), our method outperformed all others because it 1) incorporates tactile information into the leg kinematics model, and 2) adapts to changes in terrain and robot weight load based on online learning.
\textit{FAST-LIO2} failed due to the featureless areas (e.g., \fig{fig:jobutsu_traj}(b-1) and \fig{fig:sunahama_traj}(b)).
In \textit{ours w/o online learning}, the translational scales were particularly incorrect on the grass of the campus (\fig{fig:jobutsu_traj}(a)) and the sandy beach (\fig{fig:sunahama_traj}(a)). 
We consider that this learning model cannot adapt to changes in the robot's weight load.
Specifically, the robot's acceleration and velocity, expressed by dividing the reaction force by the robot's mass, are not properly described in this model.
While \textit{Unitree odometry} accurately estimated the rotation motions, its translational scales were incorrect, as shown in \fig{fig:jobutsu_traj}(a).
Similarly, the trajectory of \textit{Unitree odometry w/ LIO} (fusing the above Unitree odometry-based motion constraints and LiDAR-IMU constraints) has also large errors in the translational elements, as seen in \fig{fig:jobutsu_traj}(a).

\textit{Ours w/o tactile information} achieves nearly the same accuracy (ATE: \SI{0.12}{m}; RTE: \SI{0.12}{m}) as the proposed method on the sandy beach, likely because the learning model without tactile information was sufficiently trained online to capture terrain-dependent features, similar to~\cite{okawara2024neuralwheel,wisth2022vilens} that implicitly correct terrain-dependent errors by exteroceptive sensors' constraints without tactile information.
Whereas, in the campus sequence, the ATE for \textit{ours w/o tactile information} (\SI{0.63}{m}) is approximately twice as large as that of the proposed method.
Moreover, its translational scales are particularly inconsistent in the grass, as shown in \fig{fig:jobutsu_traj}(a).
We consider the reason to be that point cloud degeneration and terrain condition changes jointly occurred when the robot transitioned from gravel to grass.
In such conditions, while kinematics models without tactile information (e.g., \textit{ours w/o tactile information}, related works~\cite{okawara2024neuralwheel,wisth2022vilens}) cannot properly describe terrain-dependent features, our model can partly represent them thanks to the inclusion of tactile information.

\hl{According to the results, we quantitatively demonstrated that incorporating online learning and tactile information in the neural leg kinematics model is effective for robust odometry estimation against challenging scenarios.}

\begin{figure}[tb]
  \centering
  \includegraphics[width=1.0\linewidth]{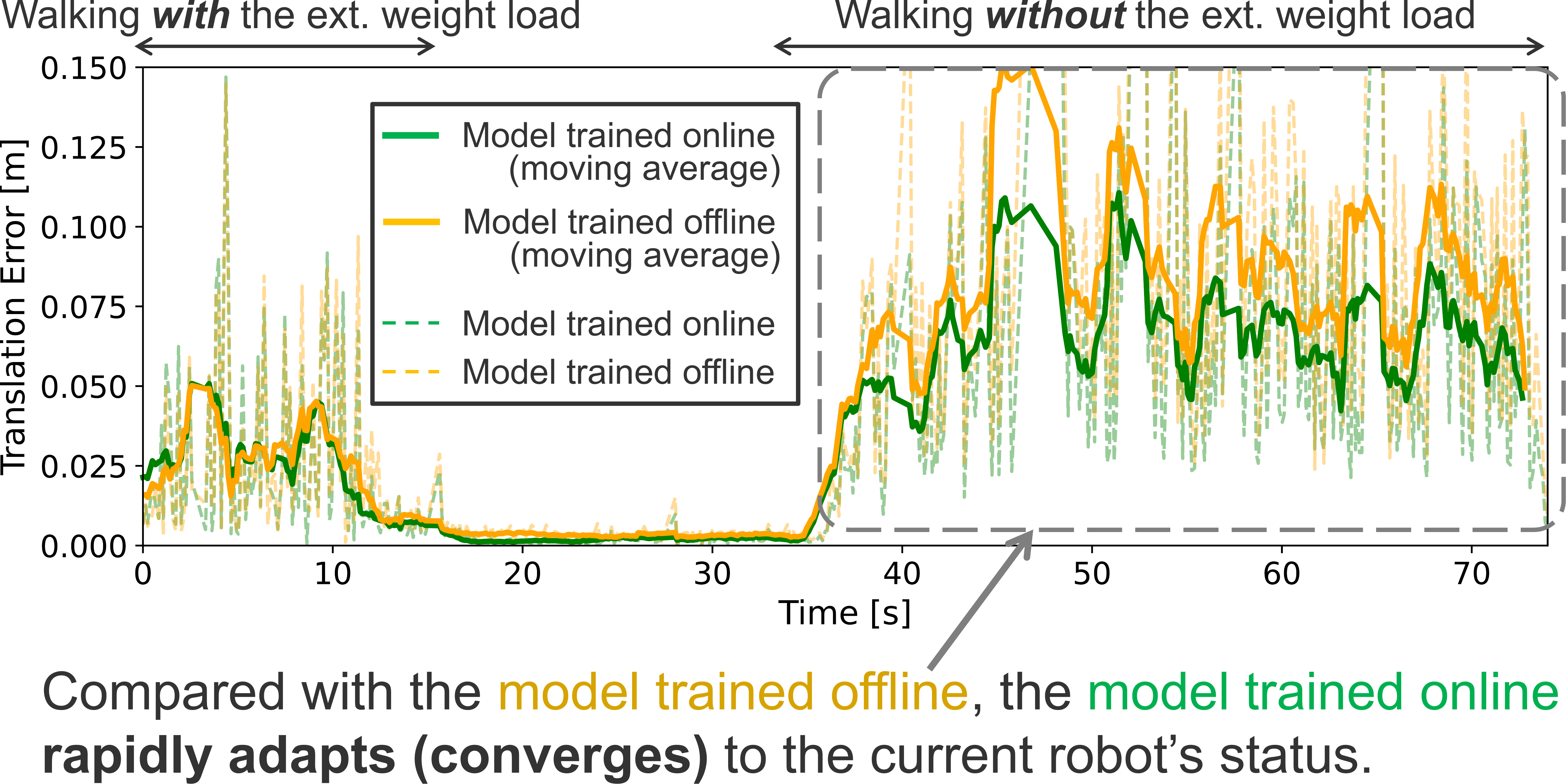}
  \caption{\hl{Time histories of motion errors for our network without LiDAR and IMU factors, during the sandy beach~(\mbox{\fig{fig:sunahama_traj}}).}}
  \label{fig:standalone_network}
\end{figure}

\hl{\textbf{The proposed network performance: }
Next, we demonstrate the performance of the neural leg kinematics model independently (i.e., LiDAR-based and IMU-based factors are not incorporated).
\mbox{\fig{fig:standalone_network}} presents the time histories of the motion errors and their moving averages for our networks trained online and offline, respectively, in the sandy beach sequence.
The models trained online and offline correspond to the kinematics model used in \textit{Ours} and \textit{Ours w/o online learning} of~\mbox{\fig{fig:sunahama_traj}}, respectively.
The online trained model shows a significant error reduction and outperforms the offline trained model thanks to its adaptability, particularly after approximately 40 seconds, when \mbox{the robot’s weight load was removed.}
While the errors slightly fluctuate due to terrain conditions and walking speed, the overall trend confirms that the proposed model adapts (converges) quickly and consistently improves its accuracy.}

\begin{figure}[tb]
  \centering
  \includegraphics[width=1.0\linewidth]{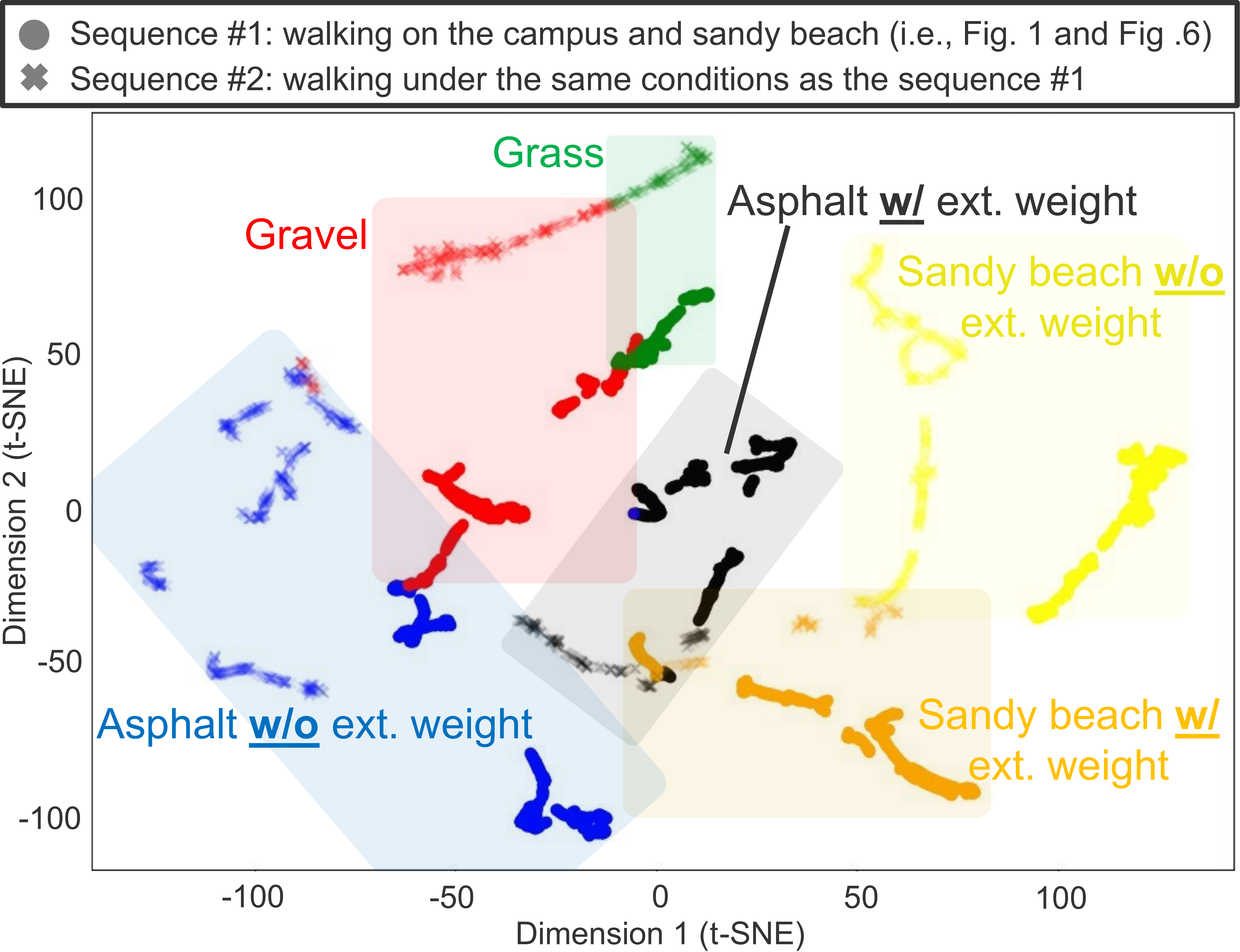}
  \caption{t-SNE-based visualization of time histories of $\bm{m}_{\text{on}}\in~\mathbb{R}^{168}$ (all weights and biases of online learning model) \hl{under the similar two sequences}.}
  \label{fig:tsne_of_online_learning_model}
\end{figure}

\textbf{Time transition of the neural leg kinematics model via online learning: }
Finally, we demonstrate that the \textit{neural leg kinematics model} was dynamically trained to adapt to terrain and the robot weight load conditions.
We analyzed a time series of the MLP parameters $\bm{m}_{\text{on}}\in \mathbb{R}^{168}$ (all weights and biases of online learning model) on the campus and the sandy beach sequence.
To aid visualization, we embedded the time series of $\bm{m}_{\text{on}}$ to 2 dimension vectors based on t-SNE~\cite{van2008visualizing}.
\fig{fig:tsne_of_online_learning_model} illustrates the 2D embedded parameter vectors.
\hl{Note that \mbox{\fig{fig:tsne_of_online_learning_model}} shows the embedded vectors for two types of sequences: 
sequence \#1: walking on the campus and sandy beach (i.e., \mbox{\fig{fig:jobutsu_traj}} and \mbox{\fig{fig:sunahama_traj}}), and 
sequence \#2: walking under the same conditions as sequence \#1.
In both sequences \#1 and \#2, we can see that $\bm{m}_{\text{on}}$ for each sequence changed similarly to adapt to the current robot situation over time.
This result indicates that our network was consistently trained online and converged to similar parameters across different sessions.
Therefore, we demonstrated that the neural leg kinematics model successfully adapts to varying robot weight loads and terrain types.}

\section{Conclusion}
We presented an odometry estimation algorithm that fuses LiDAR-IMU constraints and online trainable leg kinematic constraints incorporating tactile information (\textit{neural leg kinematics model}) in a tightly coupled way.
To effectively incorporate tactile information, the neural leg kinematics model was trained online to dynamically adapt the current robot's weight loads and terrain conditions.
We demonstrated that our odometry estimation algorithm outperformed state-of-the-art approaches (Unitree proprietary odometry with LiDAR-IMU constraints, and FAST-LIO2) and ablation study cases (our learning model without tactile information or online learning) even under various conditions, including different terrain types (asphalt, gravel, grass, and sandy beach) and robot weights (\SI{3}{kg} changes), thanks to the \textit{neural adaptive leg odometry factor} and its online uncertainty estimation.

In future work, we will extend our network to infer terrain classification, external forces acting on a robot's body and feet, in addition to the twist and foot contacts. This extension enables us to fully apply this network to legged robot control. \hl{We plan to extend our network into a more general model that accommodates various link parameters and release its code.}

\balance

\bibliographystyle{IEEEtran}
\bibliography{preprint}

\begin{thebibliography}{10}
\providecommand{\url}[1]{#1}
\csname url@samestyle\endcsname
\providecommand{\newblock}{\relax}
\providecommand{\bibinfo}[2]{#2}
\providecommand{\BIBentrySTDinterwordspacing}{\spaceskip=0pt\relax}
\providecommand{\BIBentryALTinterwordstretchfactor}{4}
\providecommand{\BIBentryALTinterwordspacing}{\spaceskip=\fontdimen2\font plus
\BIBentryALTinterwordstretchfactor\fontdimen3\font minus \fontdimen4\font\relax}
\providecommand{\BIBforeignlanguage}[2]{{%
\expandafter\ifx\csname l@#1\endcsname\relax
\typeout{** WARNING: IEEEtran.bst: No hyphenation pattern has been}%
\typeout{** loaded for the language `#1'. Using the pattern for}%
\typeout{** the default language instead.}%
\else
\language=\csname l@#1\endcsname
\fi
#2}}
\providecommand{\BIBdecl}{\relax}
\BIBdecl

\bibitem{xu2022fast}
W.~Xu, Y.~Cai, D.~He, J.~Lin, and F.~Zhang, ``{FAST-LIO2}: Fast direct {LiDAR}-inertial odometry,'' \emph{IEEE Transactions on Robotics}, vol.~38, no.~4, pp. 2053--2073, 2022.

\bibitem{okawara2024neuralwheel}
T.~Okawara, K.~Koide, S.~Oishi, M.~Yokozuka, A.~Banno, K.~Uno, and K.~Yoshida, ``Tightly-coupled {LiDAR}-{IMU}-wheel odometry with an online neural kinematic model learning via factor graph optimization,'' \emph{Robotics and Autonomous Systems}, vol. 187, no. 104929, 2025.

\bibitem{wisth2022vilens}
D.~Wisth, M.~Camurri, and M.~Fallon, ``{VILENS}: Visual, inertial, lidar, and leg odometry for all-terrain legged robots,'' \emph{IEEE Transactions on Robotics}, vol.~39, no.~1, pp. 309--326, 2022.

\bibitem{yang2022online}
S.~Yang, H.~Choset, and Z.~Manchester, ``Online kinematic calibration for legged robots,'' \emph{IEEE Robotics and Automation Letters}, vol.~7, no.~3, pp. 8178--8185, 2022.

\bibitem{vanderkop2022novel}
A.~Vanderkop, N.~Kottege, T.~Peynot, and P.~Corke, ``A novel model of interaction dynamics between legged robots and deformable terrain,'' in \emph{International Conference on Robotics and Automation}, 2022, pp. 6635--6641.

\bibitem{bloesch2013state}
M.~Bloesch, M.~Hutter, M.~A. Hoepflinger, S.~Leutenegger, C.~Gehring, C.~D. Remy, and R.~Siegwart, ``State estimation for legged robots: Consistent fusion of leg kinematics and {IMU},'' \emph{Robotics: Science and Systems VIII}, pp. 17--24, 2013.

\bibitem{hartley2018hybrid}
R.~Hartley, M.~G. Jadidi, L.~Gan, J.-K. Huang, J.~W. Grizzle, and R.~M. Eustice, ``Hybrid contact preintegration for visual-inertial-contact state estimation using factor graphs,'' in \emph{International Conference on Intelligent Robots and Systems}, 2018, pp. 3783--3790.

\bibitem{wisth2019robust}
D.~Wisth, M.~Camurri, and M.~Fallon, ``Robust legged robot state estimation using factor graph optimization,'' \emph{IEEE Robotics and Automation Letters}, vol.~4, no.~4, pp. 4507--4514, 2019.

\bibitem{fourmy2021contact}
M.~Fourmy, T.~Flayols, P.-A. L{\'e}ziart, N.~Mansard, and J.~Sol{\`a}, ``Contact forces preintegration for estimation in legged robotics using factor graphs,'' in \emph{International Conference on Robotics and Automation}, 2021, pp. 1372--1378.

\bibitem{kang2023view}
J.~Kang, H.~Kim, and K.-S. Kim, ``{VIEW}: Visual-inertial external wrench estimator for legged robot,'' \emph{IEEE Robotics and Automation Letters}, pp. 8366--8377, 2023.

\bibitem{buchanan2022learning}
R.~Buchanan, M.~Camurri, F.~Dellaert, and M.~Fallon, ``Learning inertial odometry for dynamic legged robot state estimation,'' in \emph{Conference on robot learning}, 2022, pp. 1575--1584.

\bibitem{wassermanlegolas}
J.~Wasserman, A.~Agarwal, R.~Jangir, G.~Chowdhary, D.~Pathak, and A.~Gupta, ``Legolas: Deep leg-inertial odometry,'' in \emph{Conference on Robot Learning}, 2024.

\bibitem{yang2024state}
S.~Yang, Q.~Yang, R.~Zhu, Z.~Zhang, C.~Li, and H.~Liu, ``State estimation of hydraulic quadruped robots using invariant-{EKF} and kinematics with neural networks,'' \emph{Neural Computing and Applications}, vol.~36, no.~5, pp. 2231--2244, 2023.

\bibitem{hu2017contact}
J.~Hu and R.~Xiong, ``Contact force estimation for robot manipulator using semiparametric model and disturbance kalman filter,'' \emph{IEEE Transactions on Industrial Electronics}, vol.~65, no.~4, pp. 3365--3375, 2017.

\bibitem{KingBa15}
D.~P. Kingma, ``Adam: A method for stochastic optimization,'' \emph{arXiv preprint arXiv:1412.6980}, 2014.

\bibitem{koide2021globally}
K.~Koide, M.~Yokozuka, S.~Oishi, and A.~Banno, ``Globally consistent {3D} {LiDAR} mapping with {GPU}-accelerated {GICP} matching cost factors,'' \emph{IEEE Robotics and Automation Letters}, vol.~6, no.~4, pp. 8591--8598, 2021.

\bibitem{forster2016manifold}
C.~Forster, L.~Carlone, F.~Dellaert, and D.~Scaramuzza, ``On-manifold preintegration for real-time visual--inertial odometry,'' \emph{IEEE Transactions on Robotics}, vol.~33, no.~1, pp. 1--21, 2016.

\bibitem{kaess2012isam2}
M.~Kaess, H.~Johannsson, R.~Roberts, V.~Ila, J.~J. Leonard, and F.~Dellaert, ``isam2: Incremental smoothing and mapping using the bayes tree,'' \emph{The International Journal of Robotics Research}, vol.~31, no.~2, pp. 216--235, 2012.

\bibitem{zhang2018tutorial}
Z.~Zhang and D.~Scaramuzza, ``A tutorial on quantitative trajectory evaluation for visual (-inertial) odometry,'' in \emph{International Conference on Intelligent Robots and Systems}, 2018, pp. 7244--7251.

\bibitem{van2008visualizing}
L.~Van~der Maaten and G.~Hinton, ``Visualizing data using t-{SNE}.'' \emph{Journal of machine learning research}, vol.~9, no.~11, pp. 2579--2605, 2008.

\end{thebibliography}

\end{document}